\documentclass[letterpaper]{article} 
\usepackage[preprint]{aaai2027}  
\usepackage[hyphens]{url}  
\usepackage{graphicx} 
\usepackage{natbib}  
\usepackage{caption} 
\usepackage{algorithm}
\usepackage{algorithmic}
\usepackage{amsmath}
\usepackage{amssymb}
\usepackage{mathtools}

\usepackage{graphicx}

\usepackage{booktabs}
\usepackage{multirow}
\usepackage{array}
\usepackage{tabularx}
\usepackage[table]{xcolor}

\usepackage{xspace}
\usepackage{xcolor}

\usepackage{newfloat}
\usepackage{listings}
\DeclareCaptionStyle{ruled}{labelfont=normalfont,labelsep=colon,strut=off} 
\floatstyle{ruled}
\newfloat{listing}{tb}{lst}{}
\floatname{listing}{Listing}

\usepackage{booktabs}

\title{Distill What RGB Can Recover: Privileged 3D Evidence for RGB-Only Vision-Language Models}

\author{
    Yanbin Hu\textsuperscript{\rm 1,\rm 3}\equalcontrib,
    Jin Cui\textsuperscript{\rm 2,\rm 3}\equalcontrib,
    Jun Ye\textsuperscript{\rm 1,\rm 3},
    Jiepeng Zhou\textsuperscript{\rm 4},\\
    Jiangcheng Song\textsuperscript{\rm 2,\rm 3},
    Boran Zhao\textsuperscript{\rm 1,\rm 3}\corresponding,
    Pengju Ren\textsuperscript{\rm 2,\rm 3}
}

\affiliations{
    \textsuperscript{\rm 1}School of Software, Xi'an Jiaotong University\\
    \textsuperscript{\rm 2}School of Artificial Intelligence, Xi'an Jiaotong University\\
    \textsuperscript{\rm 3}State Key Laboratory of Human-Machine Hybrid Augmented Intelligence,\\
    Institute of Artificial Intelligence and Robotics, Xi'an Jiaotong University\\
    \textsuperscript{\rm 4}The Hong Kong University of Science and Technology (Guangzhou)\\
    yanbinhu@stu.xjtu.edu.cn; \{boranzhao, pengjuren\}@xjtu.edu.cn
}

\begin{document}

\maketitle

\begin{abstract}


3D scene understanding requires reasoning about entity existence, spatial layout, and object relations, yet RGB images alone often provide insufficient 3D cues. Existing 3D-VLMs commonly rely on depth or 3D-position-aware inputs at inference time, introducing additional acquisition, reconstruction, or annotation costs that limit RGB-only deployment. We therefore study how training-time 3D evidence can be converted into spatial reasoning capabilities retained under RGB-only inference. We propose a \textbf{privileged-evidence distillation framework} that constructs a distillable teacher through a unified evidence interface and controlled residual injection, and transfers its knowledge to a deployable student receiving only RGB images and questions through logit and structured representation distillation. To avoid imitating teacher signals unsupported by RGB, we further introduce \emph{evidence-sensitivity-guided distillation}, which uses corrupted evidence to identify highly evidence-dependent targets and down-weight their supervision. We also define a recoverability decomposition based on the matched baseline, teacher, and student, separating privileged gains into RGB-recoverable improvements and residual teacher advantages. Across four benchmarks, the teacher achieves the best result on \textbf{7 of 11 reported metrics} among the compared methods. The RGB-only student outperforms its matched baseline on \textbf{all 11 metrics}, including gains of 10.4 ScanQA CIDEr and 19.1 Scan2Cap CIDEr@0.5, without additional inference-time inputs. These results validate the effectiveness of training-time privileged 3D evidence distillation for both teacher performance and deployable RGB-only spatial reasoning. Separately, our matched baseline--teacher--student analysis characterizes privileged-gain transfer across evidence types and spatial skills.

\end{abstract}


\section{Introduction}

\begin{figure}[!t]
\centering
\includegraphics[width=\columnwidth]{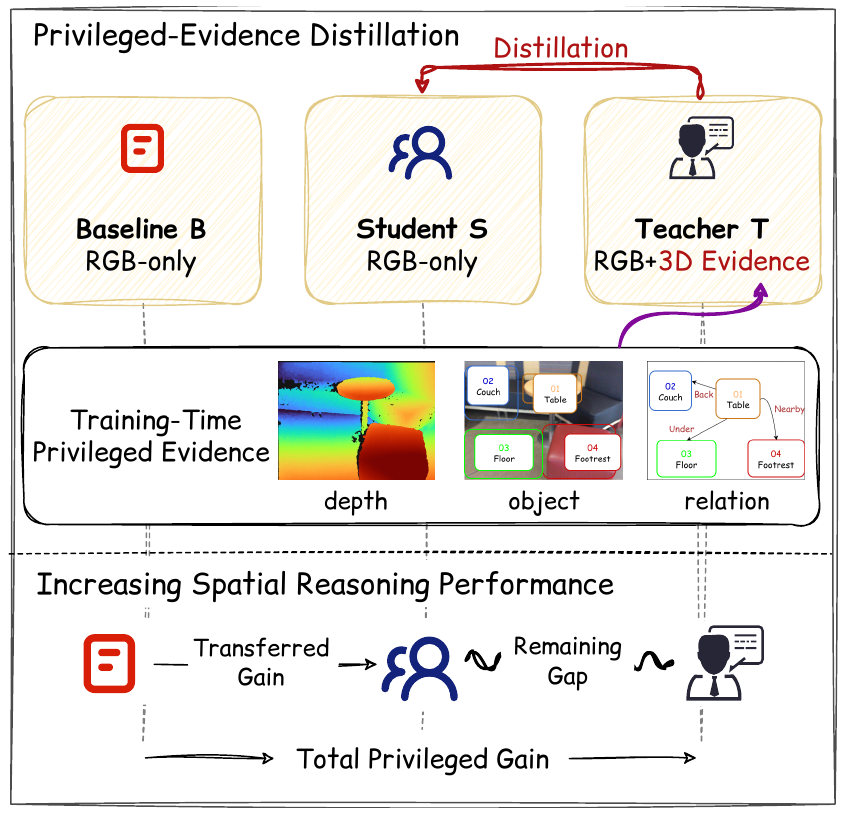}
\caption{\textbf{Conceptual overview of privileged-evidence distillation.} The teacher (T) uses training-time depth, object, and relation evidence, while the student (S) learns transferable spatial reasoning from RGB only. Relative to the matched RGB-only baseline (B), the student recovers part of the privileged gain, while the remaining gap represents knowledge not transferred to RGB-only inference.}
\label{fig:concept}
\vspace{-2mm}
\end{figure}

Multimodal large language models have made substantial progress in vision-language understanding~\cite{liu2023visual,li2023blip}, yet visual question answering in real-world 3D scenes still requires reasoning about depth, spatial layouts, object locations, and inter-object relations~\cite{azuma2022scanqa,ma2022sqa3d}. Tasks in 3D question answering, visual grounding, and embodied navigation show that spatially grounded answers require jointly interpreting appearance, scene geometry, object locations, directional relations, and viewer- or agent-centric context~\cite{azuma2022scanqa,chen2020scanrefer,ma2022sqa3d,anderson2018vision}. RGB-only VLMs provide rich appearance, semantic, and contextual cues, but observe 3D structure only through 2D projections, making depth ordering, local relations, and scene-level layouts difficult to model reliably~\cite{hong20233d,zhu2025llava}.

To improve 3D scene understanding, recent studies introduce point clouds, 3D position embeddings, or structured geometric priors during training or inference~\cite{hong20233d,zhu2025llava,chen2024ll3da}. These signals provide direct geometric and relational constraints for spatial cues that are implicit and unstable in RGB, improving 3D question answering, grounding, and reasoning. However, inference-time dependence on point clouds, reconstructed scenes, depth sensors, or structured evidence restricts deployment to environments where such inputs are available. Privileged modality learning and cross-modal distillation instead allow additional modalities to supervise training while deployment relies only on the target modality~\cite{vapnik2009new,luo2018graph,dai2021learning}. A deployable RGB-only 3D-VLM should therefore exploit 3D evidence during training without requiring it at inference time.

This motivates a teacher--student paradigm in which the teacher accesses RGB and privileged 3D evidence during training, while the student receives only RGB and distills the teacher's RGB-supported spatial reasoning ability~\cite{hinton2015distilling,lopez2015unifying,garcia2018modality}. Depth, object, and relation evidence thus serve as training-time supervision rather than inference-time inputs, helping the student learn stronger scene representations~\cite{liu20213d,dai2021learning}. However, 3D gains are not uniformly recoverable from RGB: relative depth, coarse layouts, and some relations may have monocular cues, whereas metric distances or invisible relations often lack visual support. Fully imitating an evidence-dependent teacher may therefore transfer unrecoverable answer biases. The central challenge is to construct a distillable teacher and selectively transfer spatial reasoning gains supported by RGB.

To this end, we propose a privileged-evidence injection and recoverability-decomposed distillation framework for RGB-only 3D-VLMs. It contains three models with the same VLM backbone: an RGB-only baseline (B), a teacher (T) with training-time access to RGB and privileged 3D evidence, and a distilled student (S) using only RGB at inference time~\cite{hinton2015distilling,lopez2015unifying}. Through a unified type-agnostic interface, the teacher compresses geometric, object, and relation evidence into tokens and injects compact residuals into router-selected backbone layers. Motivated by findings that task-relevant information and contextual knowledge are distributed differently across Transformer depths~\cite{ciernikattentive,ju2024large}, layer-wise injection lets evidence directly update representations at selected depths. Compared with input-only prefix fusion, it reduces single-sequence RGB--evidence--question alignment difficulty and the context overhead of long object--relation prefixes. Controlled residuals further limit deviation from the original RGB representation. The student receives no 3D evidence and learns RGB-recoverable gains through logit and structural hidden-state distillation. We use corrupted evidence to estimate teacher dependence and down-weight highly evidence-sensitive samples, reducing the transfer of unrecoverable information. Comparing (T), (S), and (B) then reveals recoverable gains and remaining residuals across evidence types and spatial reasoning skills. Our contributions are summarized as follows:

\begin{itemize}

\item \textbf{Privileged-evidence distillation for RGB-only 3D-VLMs.}
We study a deployment-oriented setting where 3D evidence is available during training but absent at inference, and introduce a framework that leverages geometric, object, and relational evidence without additional 3D inputs or processing during deployment.

\item \textbf{Injection--distillation co-design for selective knowledge transfer.}
A unified evidence interface and controlled residual injection construct a strong yet distillable teacher, while logit distillation, structural representation transfer, and counterfactual evidence-sensitivity weighting selectively transfer RGB-supported spatial reasoning.

\item \textbf{Privileged-gain transfer analysis.}
We quantify transferred gains and remaining residuals through the performance relationships among a matched RGB baseline, privileged teacher, and distilled student. Across multiple 3D scene understanding benchmarks, the privileged teacher achieves leading or competitive performance on a broad range of metrics, while the RGB-only student consistently outperforms its matched baseline.

\end{itemize}

\section{Related Work}

\paragraph{3D vision-language understanding.}
3D vision-language understanding is motivated by practical demands for object grounding, referring expression comprehension, question answering, and spatial relation reasoning in 3D scenes. Compared with 2D image understanding, it requires models not only to recognize object categories, but also to model object existence, viewpoint changes, spatial layouts, relative positions, and inter-object relations. With the increasing reasoning and instruction-following abilities of large language models, recent methods further connect 3D scene representations with LLMs. For example, 3D-LLM, LLaVA-3D, and 3D-LLaVA exploit point clouds, 3D position embeddings, scene tokens, or superpoint representations to enhance 3D dialogue and spatial reasoning~\cite{hong20233d,zhu2025llava,deng20253d}. However, these methods usually require explicit 3D inputs, 3D encoders, or reconstructed scene representations at inference time, introducing additional acquisition, reconstruction, and alignment costs when only RGB images are available. Since RGB images still contain perspective, occlusion, relative scale, and semantic context cues that partially reveal spatial structure, training-time 3D supervision can serve as an auxiliary signal for improving RGB-based spatial representations without necessarily retaining 3D inputs during deployment.

\paragraph{Geometry-aware VLMs.}
Recent studies also improve spatial reasoning under standard VLMs or lighter 3D-input settings. SpatialRGPT enhances region-level spatial perception and reasoning by curating regional data from 3D scene graphs and introducing a depth-aware plugin module~\cite{cheng2024spatialrgpt}. VLM-3R targets monocular videos, deriving spatial tokens and view tokens with a geometry encoder and using 3D reconstructive instruction tuning to improve spatial-temporal reasoning~\cite{fan2026vlm}. GASP injects geometric priors into VLM transformer layers through point-correspondence and depth-consistency supervision~\cite{yeh2026beyond}. SpatialStack synchronizes multi-level geometric features with the language backbone to alleviate the bottleneck of fusing only deep-layer geometry features~\cite{zhang2026spatialstack}. These methods indicate that VLM spatial ability can be improved through geometric supervision, implicit 3D representations, or hierarchical fusion.

\paragraph{Privileged evidence distillation.}
Learning using privileged information and knowledge distillation provide related foundations for using additional evidence during training while keeping model inputs constrained at inference time~\cite{vapnik2009new,hinton2015distilling}. In 3D scene understanding, depth, object annotations, and relation graphs can provide more direct spatial evidence than RGB images, but such evidence is often unavailable in standard RGB-only deployment. Therefore, the relevant problem is not merely to compress a larger teacher, but to transfer the benefits induced by training-time 3D evidence to an RGB-only student, while avoiding forcing the student to imitate teacher behaviors that are unrecoverable from RGB inputs. This makes teacher construction, distillation targets, and sample-level evidence dependency important factors in privileged distillation for 3D understanding.

\section{Method}
\label{sec:method}

\begin{figure*}[t]
    \centering
    \includegraphics[width=\textwidth]{"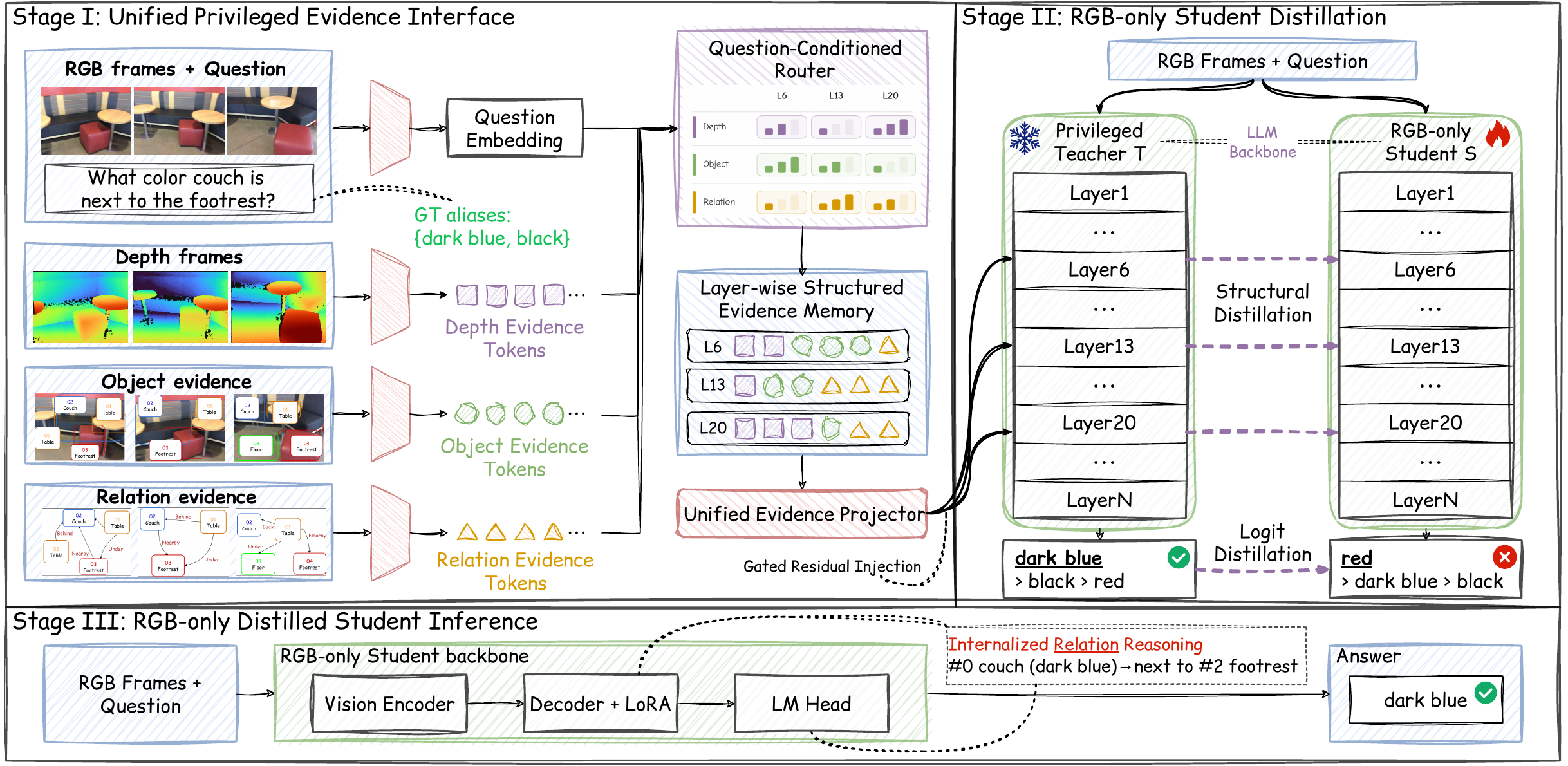"}
    \caption{\textbf{Overview of our privileged-evidence distillation framework.} During training, depth, object, and relation evidence is fused into Unified Evidence Tokens and injected into selected language layers to construct the privileged teacher \(T\). The RGB-only student \(S\) learns from the teacher through logit and structural distillation while retaining the original RGB-only input pathway. At inference, all evidence modules are removed and the student uses only RGB frames and the question. Layer indices are illustrated using LLaVA-Video and can be adapted to other backbones.}
    \label{fig:framework}
\end{figure*}

\subsection{Problem Formulation and Framework Overview}
\label{sec:problem_formulation}

Given an RGB input $I$, a question $q$, and a target answer $y$, we assume access to privileged 3D evidence $\mathcal{E}=\{D,O,R\}$ during training, where $D$, $O$, and $R$ denote depth, object, and relation evidence. Our framework contains an RGB-only baseline $B$, a privileged teacher $T$, and an RGB-only distilled student $S$, all built on the same vision-language architecture:
\begin{equation}
\begin{aligned}
p_B(y\mid I,q),\quad
p_T(y\mid I,q,\mathcal{E}),\quad
p_S(y\mid I,q).
\end{aligned}
\label{eq:problem_formulation}
\end{equation}

The baseline and student share the same inputs, backbone, and trainable capacity, while only the student receives teacher supervision. The teacher incorporates privileged evidence through modality encoders, a question-conditioned router, a unified projector, and sparse residual injection. Our goal is to construct a strong yet distillable teacher and transfer the gains supported by RGB cues to the student.

Training proceeds through two teacher phases and one student-distillation phase. Phase A freezes the teacher LoRA and learns the evidence interface; Phase B freezes the evidence encoders and adapts the teacher LoRA and injection modules. The frozen teacher then supervises the student. The baseline is trained independently with the same initialization, data, trainable-parameter configuration, and optimization steps as the student.

\subsection{Unified Privileged Evidence Interface}
\label{sec:unified_evidence}

For each scene, we use an RGB-aligned depth map, cached object descriptors, and pairwise relations computed from selected objects, which respectively encode geometry, entity semantics, and inter-object structure. For video inputs, the depth map corresponds to the middle sampled frame. Lightweight modality-specific encoders map the evidence into a shared space:
\begin{equation}
\begin{aligned}
Z_d = f_d(D), \quad
Z_o = f_o(O), \quad
Z_r = f_r(R),
\end{aligned}
\label{eq:evidence_encoding}
\end{equation}
where $D$, $O$, and $R$ denote the depth, object, and relation inputs.

Rather than directly concatenating the three streams, a question-conditioned router predicts their contributions to $L_v$ virtual fusion slots:
\begin{equation}
\begin{aligned}
A=
\operatorname{Softmax}
\left(
f_{\mathrm{route}}
\left(
[\bar{q};\bar{Z}_d;\bar{Z}_o;\bar{Z}_r]
\right)
\right),
A \in \mathbb{R}^{L_v \times 3},
\end{aligned}
\label{eq:evidence_routing}
\end{equation}
and the routed features are aggregated as
\begin{equation}
F =
\operatorname{ModalityFusion}
\left(
Z_d,Z_o,Z_r;A
\right),
F\in\mathbb{R}^{L_v\times d_e},
\label{eq:evidence_fusion}
\end{equation}
where $d_e$ is the shared evidence dimension.

Learnable queries extract a fixed number of \emph{Unified Evidence Tokens} through Q-Former-style cross-attention:
\begin{equation}
\begin{aligned}
U
=
W_p\,
\operatorname{CrossAttn}
\left(
Q_u,F,F
\right),
\quad
U \in\mathbb{R}^{N_u\times d_L},
\end{aligned}
\label{eq:unified_evidence_tokens}
\end{equation}
where $Q_u$ contains $N_u$ learnable queries, $W_p$ projects to the language feature space, and $d_L$ is the language hidden dimension.

Instead of prepending $U$ to the input sequence, we expose it to a sparse set of language layers through controlled residual cross-attention:
\begin{equation}
H_k^{T}
=
H_k
+
\gamma_k
\operatorname{CrossAttn}
\left(
\operatorname{LN}(H_k), U, U
\right), k \in \mathcal{S}_L,
\label{eq:unified_evidence_injection}
\end{equation}
where $\mathcal{S}_L$ is a backbone-specific injection set and $\gamma_k$ controls evidence strength. This sparse multi-layer design lets different reasoning stages access shared evidence and adapts naturally to backbones of different depths.

\subsection{Distillable Privileged Teacher}
\label{sec:distillable_teacher}

The teacher is used only during training. Its predictions and intermediate representations supervise the student after evidence injection. Small-initialized residual scales and low-probability evidence dropout encourage evidence use while preserving the original evidence-free RGB pathway.

\paragraph{Teacher Phase A: Evidence-Interface Learning.}
We freeze the teacher LoRA and optimize the evidence encoders, router, unified projector, and residual injection modules:
\begin{equation}
\mathcal{L}_{T}^{A}
=
\mathcal{L}_{\mathrm{task}}^{T}
+
\lambda_{\mathrm{depth}}
\mathcal{L}_{\mathrm{depth}},
\label{eq:teacher_stage_a}
\end{equation}
where $\mathcal{L}_{\mathrm{task}}^{T}$ is autoregressive cross-entropy. The auxiliary depth loss combines value reconstruction and local gradient consistency on the encoded depth representation, does not pass through the language backbone, and is disabled after Phase A.

\paragraph{Teacher Phase B: Privileged-Teacher Adaptation.}
We freeze the evidence encoders and optimize the teacher LoRA and residual injection modules:
\begin{equation}
\mathcal{L}_{T}^{B}
=
\mathcal{L}_{\mathrm{task}}^{T}
+
\lambda_{\mathrm{evi}}
\mathcal{L}_{\mathrm{evi}},
\label{eq:teacher_stage_b}
\end{equation}
where $\mathcal{L}_{\mathrm{evi}}$ contrasts clean and corrupted evidence by favoring lower answer negative log-likelihood under clean evidence and preserving separation between their evidence representations. This discourages gains arising only from additional module capacity.

\subsection{RGB-Only Student Distillation}
\label{sec:student_distillation}

The student shares the teacher backbone but receives no Unified Evidence Tokens and disables evidence injection. After teacher training, the frozen teacher produces stop-gradient targets from an evidence-enabled forward pass, while the student receives only RGB and the question.

We distill output distributions and intermediate token-interaction structures in the language backbone, but do not apply intermediate supervision to the visual encoder to avoid simultaneously moving visual and language representations.

Let $\ell_T$ and $\ell_S$ denote teacher and student logits. Their temperature-scaled distributions are
\begin{equation}
p_T^{\tau}
=
\operatorname{Softmax}(\ell_T/\tau),
\quad
p_S^{\tau}
=
\operatorname{Softmax}(\ell_S/\tau),
\label{eq:distillation_distributions}
\end{equation}
and the logit distillation loss is
\begin{equation}
\mathcal{L}_{\mathrm{kd}}
=
\tau^2
D_{\mathrm{KL}}
\left(
\operatorname{sg}(p_T^{\tau})
\,\|\,p_S^{\tau}
\right),
\label{eq:logit_distillation}
\end{equation}
where $\operatorname{sg}(\cdot)$ denotes stop-gradient.

For structural distillation, we align pairwise token relations rather than hidden states dimension by dimension:
\begin{equation}
\Phi(H)
=
\operatorname{Softmax}
\left(
\frac{HH^{\top}}{\sqrt{d}}
\right),
\label{eq:structural_mapping}
\end{equation}
where softmax is applied row-wise. At each selected layer,
\begin{equation}
R_k^{T}
=
\Phi(H_k^{T}),
\quad
R_k^{S}
=
\Phi\!\left(A_k(H_k^{S})\right),
\label{eq:teacher_student_structure}
\end{equation}
where $A_k$ is a lightweight student adapter. The loss is
\begin{equation}
\mathcal{L}_{\mathrm{str}}
=
\frac{1}{|\mathcal{S}_D|}
\sum_{k\in\mathcal{S}_D}
D_{\mathrm{KL}}
\left(
\operatorname{sg}(R_k^{T})
\,\|\,R_k^{S}
\right),
\label{eq:structural_distillation}
\end{equation}
where $\mathcal{S}_D$ denotes the distillation layers. This transfers token-interaction structure without requiring the student to reconstruct privileged evidence.

\paragraph{Evidence-Sensitivity Weighted Distillation.}
Teacher targets that strongly depend on inaccessible evidence may be poorly transferable. We therefore compare predictions under the original evidence $\mathcal{E}$ and a counterfactually perturbed counterpart $\widetilde{\mathcal{E}}$:
\begin{equation}
\begin{aligned}
p_T^{\mathcal{E}}
&=
p_T(\cdot\mid I,q,\mathcal{E}),\\
p_T^{\widetilde{\mathcal{E}}}
&=
p_T(\cdot\mid I,q,\widetilde{\mathcal{E}}).
\end{aligned}
\label{eq:clean_corrupted_predictions}
\end{equation}

We replace depth with sample-mismatched features and independently shuffle object and relation features, approximately preserving feature scale and marginal statistics while disrupting their semantic correspondence.

The evidence sensitivity and distillation weight are
\begin{equation}
\begin{aligned}
s(x)
&=
D_{\mathrm{JS}}
\left(
p_T^{\mathcal{E}}
\,\|\,
p_T^{\widetilde{\mathcal{E}}}
\right),\\
w(x)
&=
\exp\left(-\kappa s(x)\right),
\end{aligned}
\label{eq:evidence_sensitivity}
\end{equation}
where $\kappa$ controls attenuation. Larger $s(x)$ indicates stronger dependence on privileged evidence and thus a less directly transferable target.

The complete student objective is
\begin{equation}
\mathcal{L}_{S}
=
\mathcal{L}_{\mathrm{task}}^{S}
+
w(x)
\left(
\lambda_{\mathrm{kd}}\mathcal{L}_{\mathrm{kd}}
+
\lambda_{\mathrm{str}}\mathcal{L}_{\mathrm{str}}
\right).
\label{eq:student_total_loss}
\end{equation}
The weight $w(x)$ applies only to teacher-derived supervision; the ground-truth task loss remains unweighted. The matched baseline is independently optimized using
\begin{equation}
\mathcal{L}_{B}
=
\mathcal{L}_{\mathrm{task}}^{B}.
\label{eq:baseline_loss}
\end{equation}

\paragraph{Privileged-Gain Transfer Analysis.}
For any metric $\widehat{M}$ oriented such that larger is better, the teacher gain over the matched RGB baseline decomposes as
\begin{equation}
\underbrace{
\widehat{M}(T)-\widehat{M}(B)
}_{\displaystyle \Delta_{\mathrm{priv}}}
=
\underbrace{
\widehat{M}(S)-\widehat{M}(B)
}_{\displaystyle \Delta_{\mathrm{trans}}}
+
\underbrace{
\widehat{M}(T)-\widehat{M}(S)
}_{\displaystyle \Delta_{\mathrm{res}}}.
\label{eq:privileged_gain_decomposition}
\end{equation}
Here, $\Delta_{\mathrm{priv}}$ is the total privileged gain, $\Delta_{\mathrm{trans}}$ is the portion retained by the student, and $\Delta_{\mathrm{res}}$ is the remaining teacher--student gap.

We define the privileged-gain transfer ratio and normalized residual as
\begin{equation}
\begin{aligned}
\mathcal{R}_{M}
&=
\frac{
\widehat{M}(S)-\widehat{M}(B)
}{
\widehat{M}(T)-\widehat{M}(B)
}
=
\frac{\Delta_{\mathrm{trans}}}{\Delta_{\mathrm{priv}}}, \\
\mathcal{U}_{M}
&=
1-\mathcal{R}_{M}
=
\frac{\Delta_{\mathrm{res}}}{\Delta_{\mathrm{priv}}},
\label{eq:privileged_gain_transfer}
\end{aligned}
\end{equation}
where $\mathcal{R}_{M}$ measures the retained teacher gain and $\mathcal{U}_{M}$ the normalized residual. We report them only when $\Delta_{\mathrm{priv}}>\epsilon$ to exclude negligible teacher improvements.

We apply this analysis across geometry, object, relation, and combined evidence configurations, as well as relative depth, directional relations, object existence, counting, and metric distance. It separates the original evidence benefit, the portion retained by the RGB-only student, and the remaining capability gap.

\section{Experiments}
\label{sec:experiments}

\subsection{Experimental Setup}
\label{sec:experimental_setup}

\paragraph{Benchmarks and Metrics.}
We evaluate our method on four complementary benchmarks covering 3D scene question answering, situated spatial reasoning, referring-expression grounding, and dense 3D captioning. On \textit{ScanQA}~\cite{azuma2022scanqa}, we report CIDEr, BLEU-4, METEOR, and ROUGE-L. On \textit{SQA3D}~\cite{ma2022sqa3d}, we report exact match (EM). \textit{ScanRefer}~\cite{chen2020scanrefer} is evaluated using Acc@0.25 and Acc@0.5. For \textit{Scan2Cap}~\cite{chen2021scan2cap}, we report CIDEr, BLEU-4, METEOR, and ROUGE-L at IoU@0.5.

\paragraph{Comparison Protocol.}
All of our models are initialized from the same vision-language backbone. The RGB-only baseline \(B\) and distilled student \(S\) use the same training data, model initialization, trainable-parameter budget, and number of optimization steps; \(S\) additionally receives distillation supervision from the fixed teacher \(T\). The teacher accesses depth, object, and relation evidence during training and full-model evaluation, whereas \(B\) and \(S\) receive only RGB frames and the question at inference time. The visual encoder remains frozen in all settings, and ScanRefer and Scan2Cap use task heads with the same architecture.

\paragraph{Implementation Details.}
The teacher is trained in two stages. We first optimize the evidence encoders, question-conditioned router, unified evidence projector, and residual injection modules. We then freeze the evidence encoders and jointly adapt the teacher LoRA and injection modules. During student distillation, the teacher remains fixed, while only the student LoRA and distillation-specific modules are updated. We use LLaVA-Video-7B-Qwen2~\cite{zhang2024llava} as the default backbone, compress the privileged inputs into 80 evidence tokens, and inject them at layers 6, 13, and 20. Training uses AdamW for 1,500 optimization steps with a learning rate of $2\times10^{-5}$.

To examine whether evidence injection and distillation supervision induce object-existence hallucinations, we additionally evaluate the key mechanism ablations on 3D-POPE~\cite{yang20253d}. We report F1 and the affirmative-answer ratio (Yes\%) in the main text, which characterize overall classification performance and affirmative bias, respectively. We use 3D-POPE solely as a mechanism-oriented diagnostic and not as a training objective or a primary task benchmark.

\subsection{Main Results}
\label{sec:main_results}

\paragraph{Overall Comparison.}
Table~\ref{tab:main_results} summarizes the main results on ScanQA, SQA3D, ScanRefer, and Scan2Cap. \(B\) is the independently trained RGB-only baseline, \(T\) is the privileged teacher evaluated with complete depth, object, and relation evidence, and \(S\) is the deployable RGB-only student.

\begin{table*}[t]
    \centering
    \scriptsize
    \setlength{\tabcolsep}{2.2pt}
    \renewcommand{\arraystretch}{0.88}
    \setlength{\aboverulesep}{0.20ex}
    \setlength{\belowrulesep}{0.20ex}

    \caption{
    Quantitative comparison on ScanQA, SQA3D, ScanRefer, and Scan2Cap.
    All Scan2Cap metrics are evaluated at IoU@0.5.
    ScanRefer results are reported as Acc@0.25 and Acc@0.5.
    Underlining denotes the best overall result, while bold denotes
    the best result among our variants.
    }
    \label{tab:main_results}

    \resizebox{\textwidth}{!}{
    \begin{tabular}{l cccc c cc cccc}
        \toprule
        \multirow{2}{*}{Method}
        & \multicolumn{4}{c}{ScanQA (val)}
        & \multicolumn{1}{c}{SQA3D}
        & \multicolumn{2}{c}{ScanRefer (val)}
        & \multicolumn{4}{c}{Scan2Cap (val)} \\
        \cmidrule(lr){2-5}
        \cmidrule(lr){6-6}
        \cmidrule(lr){7-8}
        \cmidrule(lr){9-12}
        & C & B-4 & M & R
        & EM
        & A@.25 & A@.5
        & C@.5 & B-4@.5 & M@.5 & R@.5 \\
        \midrule

        \multicolumn{12}{l}{
        \textbf{\textit{Task-specific models}}} \\

        ScanQA~\cite{azuma2022scanqa}
        & 64.9 & 10.1 & 13.1 & 33.3
        & 47.2
        & -- & --
        & -- & -- & -- & -- \\

        3D-VisTA~\cite{zhu20233d}
        & 69.6 & 10.4 & 13.9 & 35.7
        & 48.5
        & -- & --
        & 61.6 & 34.1 & 26.8 & 55.0 \\

        ScanRefer~\citeyearpar{chen2020scanrefer}
        & -- & -- & -- & --
        & --
        & 37.3 & 24.3
        & -- & -- & -- & -- \\

        BUTD-DETR~\cite{jain2022bottom}
        & -- & -- & -- & --
        & --
        & 52.2 & 39.8
        & -- & -- & -- & -- \\

        Scan2Cap~\cite{chen2021scan2cap}
        & -- & -- & -- & --
        & --
        & -- & --
        & 39.1 & 23.3 & 22.0 & 44.8 \\

        Vote2Cap-DETR~\cite{chen2023end}
        & -- & -- & -- & --
        & --
        & -- & --
        & 61.8 & 34.5 & 26.2 & 54.4 \\

        \midrule
        \multicolumn{12}{l}{
        \textbf{\textit{Video-input spatial models}}} \\

        LLaVA-Video-7B~\cite{zhang2024llava}
        & 88.7 & 3.1 & 17.7 & 44.6
        & 48.5
        & -- & --
        & -- & -- & -- & -- \\

        VLM-3R~\cite{fan2026vlm}
        & 101.9 & 15.5 & 19.7 & 49.1
        & 60.7
        & -- & --
        & -- & -- & -- & -- \\

        \midrule
        \multicolumn{12}{l}{
        \textbf{\textit{Generalist 3D and geometry-assisted models}}} \\

        Chat-Scene~\cite{huang2024chat}
        & 87.7 & 14.3 & 18.0 & 41.6
        & 54.6
        & 55.5 & 50.2
        & 77.2 & 36.3 & 28.0 & 58.1 \\

        LLaVA-3D~\cite{zhu2025llava}
        & 91.7 & 14.5 & 20.7 & 50.1
        & 55.6
        & 50.1 & 42.7
        & 79.2 & 41.1 & 30.2 & 63.4 \\

        3D-LLaVA~\cite{deng20253d}
        & 92.6 & 17.1 & 18.4 & 43.1
        & 54.5
        & -- & --
        & 78.8 & 36.9 & 27.1 & 57.7 \\

        Video-3D-LLM~\citeyearpar{zheng2025video}
        & 102.1 & 16.2 & 19.8 & 49.0
        & 58.6
        & 58.1 & 51.7
        & 83.8 & 41.3 & 28.9 & 62.3 \\

        3DRS~\cite{huang20263drs}
        & 104.8 & 17.2 & 20.5 & 49.8
        & 60.6
        & 62.9 & 56.1
        & 86.1 & 41.6 & 29.0 & 62.3 \\

        CVP~\cite{chen2026cvp}
        & 107.1
        & \underline{17.8}
        & 20.8
        & 50.9
        & \underline{62.3}
        & 62.0 & 55.4
        & \underline{90.5}
        & 41.7 & 28.9 & 62.2 \\

        VEGA-3D~\cite{wu2026generation}
        & 106.3 & -- & -- & --
        & 61.3
        & \underline{63.2} & 56.2
        & 83.2 & 42.2 & -- & -- \\

        PAR3D~\cite{dai2026par3d}
        & 95.7 & 15.9 & 18.9 & 45.0
        & 54.6
        & -- & --
        & 81.4 & 37.3 & 27.5 & 57.9 \\

        \midrule
        \multicolumn{12}{l}{
        \textbf{\textit{Our models}}} \\

        \rowcolor{gray!15}
        Ours-Baseline
        & 96.8 & 14.7 & 15.0 & 39.6
        & 53.9
        & 53.0 & 52.6
        & 43.4 & 37.9 & 27.3 & 66.2 \\

        \rowcolor{gray!15}
        Ours-Student
        & 107.2 & 15.1 & 18.1 & 46.1
        & 54.3
        & 55.5 & 54.3
        & 62.5 & 41.6 & 28.8 & 67.5 \\

        \rowcolor{gray!15}
        \textbf{Ours-Teacher}
        & \underline{\textbf{108.3}}
        & \textbf{16.9}
        & \underline{\textbf{21.6}}
        & \underline{\textbf{52.3}}
        & \textbf{56.4}
        & \textbf{58.9}
        & \underline{\textbf{56.5}}
        & \textbf{79.7}
        & \underline{\textbf{45.7}}
        & \underline{\textbf{30.4}}
        & \underline{\textbf{69.1}} \\

        \bottomrule
    \end{tabular}
    }
\end{table*}

The improvements of \(T\) over \(B\) quantify the benefit available from privileged evidence, whereas the improvements of \(S\) over \(B\) measure how much of this benefit is retained without 3D inputs at inference time. On ScanQA CIDEr and ScanRefer Acc@0.5, \(T\) improves over \(B\) by 11.5 and 3.9 points, respectively, showing that the unified evidence interface benefits both open-ended question answering and precise spatial grounding. On the same metrics, \(S\) reaches 107.2 and 54.3, respectively, and consistently outperforms \(B\) while preserving RGB-only inference.

Using the privileged-gain recovery, and taking ScanQA CIDEr, SQA3D EM, ScanRefer Acc@0.5, and Scan2Cap CIDEr@0.5 as the corresponding primary metrics, \(S\) retains 90.4\%, 16.0\%, 43.6\%, and 52.6\% of the teacher improvement, respectively. Recovery is highest on ScanQA and lowest on SQA3D, revealing clear task-dependent differences in how much privileged improvement can be retained by the RGB-only student.

\subsection{Ablation Studies}
\label{sec:ablation}

We study three questions directly related to the proposed design: where privileged evidence should be injected, whether evidence-sensitivity weighting improves effective student transfer, and how much each evidence branch contributes to the teacher. ScanQA serves as the shared task benchmark across ablations, with SQA3D, ScanRefer, Scan2Cap, or 3D-POPE additionally reported where relevant to the mechanism under study.

\paragraph{Evidence Injection Position.}
We compare input-prefix fusion, injection into a single intermediate layer, sparse multi-layer injection excluding the final language layer, and sparse multi-layer injection including the final layer. This experiment changes only the teacher injection positions and reports both task performance and 3D-POPE, allowing us to examine whether interventions close to the output distribution introduce an affirmative-answer bias.

\begin{table}[t]
    \centering
    \scriptsize
    \setlength{\tabcolsep}{2.5pt}
    \renewcommand{\arraystretch}{0.92}

    \caption{
    Ablation of teacher injection positions.
    Sparse multi-layer injection excluding the final layer achieves
    the best balance between task performance and hallucination robustness.
    C and EM denote ScanQA CIDEr and SQA3D exact match, respectively;
    F1 and Yes\% are measured on 3D-POPE.
    }
    \label{tab:injection_ablation}

    \resizebox{\columnwidth}{!}{
    \begin{tabular}{lcccc}
        \toprule
        \textbf{Injection strategy}
        & \textbf{ScanQA C}
        & \textbf{SQA3D EM}
        & \textbf{3D-POPE F1}
        & \textbf{Yes\%} \\
        \midrule

        Input prefix
        & 103.4
        & 54.9
        & 72.4
        & 72.9 \\

        Single intermediate layer
        & 106.2
        & 55.7
        & 76.9
        & 64.8 \\

        Sparse multi-layer, with final
        & 107.6
        & 56.1
        & 66.7
        & 100.0 \\

        \rowcolor{gray!15}
        \textbf{Sparse multi-layer, w/o final}
        & \textbf{108.3}
        & \textbf{56.4}
        & \textbf{80.8}
        & \textbf{56.2} \\

        \bottomrule
    \end{tabular}
    }
\end{table}

Sparse multi-layer injection excluding the final language layer achieves 108.3 ScanQA CIDEr and 56.4 SQA3D EM, outperforming both input-prefix fusion and single-layer injection on both metrics. Extending injection to the final language layer does not improve either task metric and increases Yes\% to 100.0, indicating that overly late evidence intervention can directly bias the answer distribution. We therefore use sparse multi-layer injection without the final layer in subsequent experiments.

\paragraph{Evidence-Sensitivity-Guided Distillation.}
To isolate evidence-sensitivity weighting, we fix the same full teacher and train both student variants with identical task, logit-distillation, and structural-distillation objectives. The only difference is whether the distillation weight is adjusted according to the teacher's dependence on privileged evidence, avoiding confounding changes in teacher quality with changes in student supervision.

\begin{table}[t]
    \centering
    \scriptsize
    \setlength{\tabcolsep}{2.5pt}
    \renewcommand{\arraystretch}{0.92}

    \caption{
    Ablation of evidence-sensitivity weighting.
    Sensitivity weighting improves transferable-gain recovery and
    reduces affirmative-answer bias relative to uniform distillation.
    \(\overline{\mathcal{R}}\) averages recoverability on ScanQA CIDEr
    and ScanRefer Acc@0.5; F1 and Yes\% are measured on 3D-POPE.
    }
    \label{tab:sensitivity_ablation}

    \resizebox{\columnwidth}{!}{
    \begin{tabular}{lccccc}
        \toprule
        \textbf{Variant}
        & \textbf{C}
        & \textbf{A@.5}
        & \textbf{\(\overline{\mathcal{R}}\) (\%)}
        & \textbf{F1}
        & \textbf{Yes\%} \\
        \midrule

        \(B\): task supervision
        & 96.8
        & 52.6
        & --
        & 71.8
        & 69.4 \\

        \(S\): w/o sensitivity weighting
        & 104.8
        & 53.7
        & 48.9
        & 68.9
        & 83.7 \\

        \rowcolor{gray!15}
        \textbf{\(S\): sensitivity-weighted}
        & \textbf{107.2}
        & \textbf{54.3}
        & \textbf{67.0}
        & \textbf{77.6} 
        & \textbf{58.9} \\ 

        \bottomrule
    \end{tabular}
    }
\end{table}

With the teacher fixed, uniform distillation yields an average recovery of 48.9\%, with 3D-POPE F1 and Yes\% of 68.9 and 83.7. Evidence-sensitivity weighting further improves recovery to 67.0\%, increases F1 by 8.7 points, and reduces Yes\% by 24.8 percentage points. Weighting supervision by privileged-evidence dependence therefore improves task transfer while reducing affirmative-answer bias.

\paragraph{Contribution of Privileged Evidence.}
We estimate the marginal contribution of each privileged evidence type by removing the depth, object, or relation branch at evaluation time. The removed branch is replaced using the same corruption operator employed for sensitivity estimation. This analysis measures each branch's contribution within the full teacher rather than its standalone capability under separate training.

\begin{table}[t]
    \centering
    \scriptsize
    \setlength{\tabcolsep}{2.5pt}
    \renewcommand{\arraystretch}{0.92}

    \caption{
    Leave-one-evidence-out analysis reveals complementary modality roles:
    object cues contribute most to open-ended QA and captioning, relation
    cues are particularly important for situated reasoning and grounding,
    and combining all evidence yields the strongest overall performance.
    }
    \label{tab:evidence_contribution}

    \resizebox{\columnwidth}{!}{
    \begin{tabular}{lcccc}
        \toprule
        \textbf{Available evidence}
        & \textbf{ScanQA C}
        & \textbf{SQA3D EM}
        & \textbf{ScanRefer A@.5}
        & \textbf{Scan2Cap C@.5} \\
        \midrule

        \(O+R\) (w/o depth)
        & 106.8 & 55.7 & 54.9 & 71.3 \\

        \(D+R\) (w/o object)
        & 103.6 & 55.6 & 55.3 & 65.1 \\

        \(D+O\) (w/o relation)
        & 104.7 & 54.9 & 54.5 & 68.5 \\

        Evidence off
        & 97.2 & 54.1 & 53.0 & 44.8 \\

        \midrule

        RGB-only baseline \(B\)
        & 96.8 & 53.9 & 52.6 & 43.4 \\

        \rowcolor{gray!15}
        \textbf{\(D+O+R\)}
        & \textbf{108.3}
        & \textbf{56.4}
        & \textbf{56.5}
        & \textbf{79.7} \\

        \bottomrule
    \end{tabular}
    }
\end{table}

The leave-one-evidence-out results show that removing object evidence causes the largest drops on ScanQA CIDEr and Scan2Cap CIDEr@0.5, by 4.7 and 14.6 points, respectively, whereas removing relation evidence causes the largest drops on SQA3D EM and ScanRefer Acc@0.5, by 1.5 and 2.0 points. Depth evidence provides smaller but consistent complementary gains, with its clearest effect also observed on Scan2Cap, where removal reduces CIDEr@0.5 by 8.4 points. The complete evidence configuration achieves the best result on all four benchmarks, indicating that the three evidence types provide complementary information rather than a single branch accounting for the entire teacher gain.

\subsection{Evidence- and Skill-Wise Recoverability}
\label{sec:evidence_transfer}

Aggregate benchmark scores do not show whether spatial capabilities supported by different evidence types are equally recoverable. For each example, we therefore measure the change in teacher loss after removing each evidence branch and use it as the evidence-importance score for that example. We then evaluate \(B\), \(T\), and \(S\) over relative-depth, directional-relation, object-existence, counting, occlusion, and metric-distance subsets and compute privileged-gain recovery.

To form the evidence--skill matrix, each example is assigned to the evidence type with the largest positive importance score; the aggregate row includes all evaluated examples. Recovery is omitted when the teacher improvement in a cell is below \(\epsilon\), preventing unstable ratios caused by a near-zero denominator.

\begin{figure}[t]
    \centering
    \includegraphics[width=\columnwidth]{"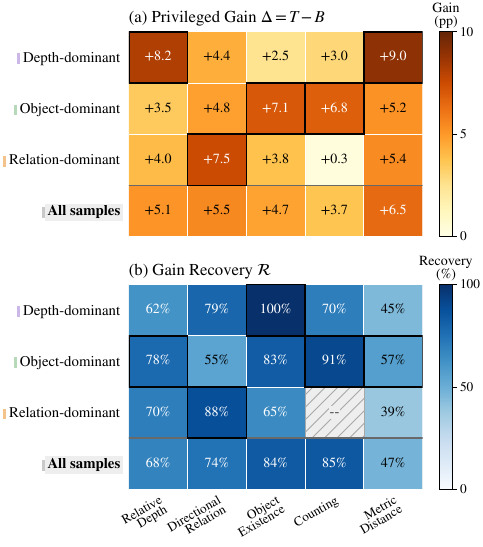"}
    \caption{\textbf{Evidence- and skill-wise privileged gain and recoverability.} The upper heatmap reports the teacher improvement over the RGB-only baseline, and the lower heatmap reports the fraction of that improvement retained by the student. Rows correspond to depth, object, relation, and aggregate evidence groups; columns correspond to spatial-skill subsets. Blank cells indicate teacher gains below \(\epsilon\).}
    \label{fig:evidence_skill_heatmap}
\end{figure}

Figure~\ref{fig:evidence_skill_heatmap} jointly presents teacher gains and student recovery across evidence--skill cells. Cells with high teacher gains but low recovery identify capabilities that remain more dependent on explicit 3D inputs, whereas cells with both high gains and high recovery indicate improvements that can be substantially internalized by the RGB-only student. This analysis distinguishes whether an evidence type is useful from whether the resulting improvement can be internalized from RGB and complements the aggregate benchmark results with a finer-grained view.
\section{Conclusion}
\label{sec:conclusion}

We present a privileged-evidence distillation framework for RGB-only 3D-VLMs. Training-time depth, object, and relation evidence is encoded as Unified Evidence Tokens and incorporated through question-conditioned residual injection, while an RGB-only student learns through logit, structural, and sensitivity-weighted distillation. Experiments show that the student consistently outperforms the matched RGB baseline while narrowing the gap to the privileged teacher. Ablations validate sparse multi-layer injection, sensitivity weighting, and complementary evidence branches, while privileged-gain analysis characterizes transfer across evidence types and spatial skills.


\begin{thebibliography}{35}
\providecommand{\natexlab}[1]{#1}

\bibitem[{Anderson et~al.(2018)Anderson, Wu, Teney, Bruce, Johnson,
  S{\"u}nderhauf, Reid, Gould, and Van Den~Hengel}]{anderson2018vision}
Anderson, P.; Wu, Q.; Teney, D.; Bruce, J.; Johnson, M.; S{\"u}nderhauf, N.;
  Reid, I.; Gould, S.; and Van Den~Hengel, A. 2018.
\newblock Vision-and-language navigation: Interpreting visually-grounded
  navigation instructions in real environments.
\newblock In \emph{Proceedings of the IEEE conference on computer vision and
  pattern recognition}, 3674--3683.

\bibitem[{Azuma et~al.(2022)Azuma, Miyanishi, Kurita, and
  Kawanabe}]{azuma2022scanqa}
Azuma, D.; Miyanishi, T.; Kurita, S.; and Kawanabe, M. 2022.
\newblock Scanqa: 3d question answering for spatial scene understanding.
\newblock In \emph{proceedings of the IEEE/CVF conference on computer vision
  and pattern recognition}, 19129--19139.

\bibitem[{Chen, Chang, and Nie{\ss}ner(2020)}]{chen2020scanrefer}
Chen, D.~Z.; Chang, A.~X.; and Nie{\ss}ner, M. 2020.
\newblock Scanrefer: 3d object localization in rgb-d scans using natural
  language.
\newblock In \emph{European conference on computer vision}, 202--221. Springer.

\bibitem[{Chen et~al.(2024)Chen, Chen, Zhang, Li, Yu, Fei, Zhu, Fan, and
  Chen}]{chen2024ll3da}
Chen, S.; Chen, X.; Zhang, C.; Li, M.; Yu, G.; Fei, H.; Zhu, H.; Fan, J.; and
  Chen, T. 2024.
\newblock Ll3da: Visual interactive instruction tuning for omni-3d
  understanding reasoning and planning.
\newblock In \emph{Proceedings of the IEEE/CVF conference on computer vision
  and pattern recognition}, 26428--26438.

\bibitem[{Chen et~al.(2023)Chen, Zhu, Chen, Lei, Yu, and Chen}]{chen2023end}
Chen, S.; Zhu, H.; Chen, X.; Lei, Y.; Yu, G.; and Chen, T. 2023.
\newblock End-to-end 3d dense captioning with vote2cap-detr.
\newblock In \emph{Proceedings of the IEEE/CVF conference on computer vision
  and pattern recognition}, 11124--11133.

\bibitem[{Chen et~al.(2021)Chen, Gholami, Nie{\ss}ner, and
  Chang}]{chen2021scan2cap}
Chen, Z.; Gholami, A.; Nie{\ss}ner, M.; and Chang, A.~X. 2021.
\newblock Scan2cap: Context-aware dense captioning in rgb-d scans.
\newblock In \emph{Proceedings of the IEEE/CVF conference on computer vision
  and pattern recognition}, 3193--3203.

\bibitem[{Chen et~al.(2026)Chen, Zhang, Xu, Xie, and Tu}]{chen2026cvp}
Chen, Z.; Zhang, X.; Xu, H.; Xie, J.; and Tu, Z. 2026.
\newblock Cvp: Central-peripheral vision-inspired multimodal model for spatial
  reasoning.
\newblock In \emph{Proceedings of the IEEE/CVF Winter Conference on
  Applications of Computer Vision}, 2295--2305.

\bibitem[{Cheng et~al.(2024)Cheng, Yin, Fu, Guo, Yang, Kautz, Wang, and
  Liu}]{cheng2024spatialrgpt}
Cheng, A.-C.; Yin, H.; Fu, Y.; Guo, Q.; Yang, R.; Kautz, J.; Wang, X.; and Liu,
  S. 2024.
\newblock Spatialrgpt: Grounded spatial reasoning in vision-language models.
\newblock \emph{Advances in Neural Information Processing Systems}, 37:
  135062--135093.

\bibitem[{Ciernik et~al.(2026)Ciernik, Morik, Thede, Eyring, Nakajima, Akata,
  and Muttenthaler}]{ciernikattentive}
Ciernik, L.; Morik, M.; Thede, L.; Eyring, L.; Nakajima, S.; Akata, Z.; and
  Muttenthaler, L. 2026.
\newblock Beyond the final layer: Attentive multilayer fusion for vision
  transformers.
\newblock \emph{arXiv preprint arXiv:2601.09322}.

\bibitem[{Dai, Das, and Bremond(2021)}]{dai2021learning}
Dai, R.; Das, S.; and Bremond, F. 2021.
\newblock Learning an augmented rgb representation with cross-modal knowledge
  distillation for action detection.
\newblock In \emph{Proceedings of the IEEE/CVF International Conference on
  Computer Vision}, 13053--13064.

\bibitem[{Dai et~al.(2026)Dai, Qu, Shen, Zhang, and Cao}]{dai2026par3d}
Dai, S.; Qu, Y.; Shen, Y.; Zhang, S.; and Cao, L. 2026.
\newblock PAR3D: A Unified 3D-MLLM with Part-Aware Representation for Scene
  Understanding.
\newblock \emph{arXiv preprint arXiv:2606.06485}.

\bibitem[{Deng et~al.(2025)Deng, He, Jiang, Wang, Dayoub, and
  Reid}]{deng20253d}
Deng, J.; He, T.; Jiang, L.; Wang, T.; Dayoub, F.; and Reid, I. 2025.
\newblock 3d-llava: Towards generalist 3d lmms with omni superpoint
  transformer.
\newblock In \emph{Proceedings of the Computer Vision and Pattern Recognition
  Conference}, 3772--3782.

\bibitem[{Fan et~al.(2026)Fan, Zhang, Li, Zhang, Chen, Hu, Wang, Wang, Qu, Zhou
  et~al.}]{fan2026vlm}
Fan, Z.; Zhang, J.; Li, R.; Zhang, J.; Chen, R.; Hu, H.; Wang, K.; Wang, P.;
  Qu, H.; Zhou, S.; et~al. 2026.
\newblock Vlm-3r: Vision-language models augmented with instruction-aligned 3d
  reconstruction.
\newblock In \emph{Proceedings of the IEEE/CVF Conference on Computer Vision
  and Pattern Recognition}, 31054--31065.

\bibitem[{Garcia, Morerio, and Murino(2018)}]{garcia2018modality}
Garcia, N.~C.; Morerio, P.; and Murino, V. 2018.
\newblock Modality distillation with multiple stream networks for action
  recognition.
\newblock In \emph{Proceedings of the European Conference on Computer Vision
  (ECCV)}, 103--118.

\bibitem[{Hinton, Vinyals, and Dean(2015)}]{hinton2015distilling}
Hinton, G.; Vinyals, O.; and Dean, J. 2015.
\newblock Distilling the knowledge in a neural network.
\newblock \emph{arXiv preprint arXiv:1503.02531}.

\bibitem[{Hong et~al.(2023)Hong, Zhen, Chen, Zheng, Du, Chen, and
  Gan}]{hong20233d}
Hong, Y.; Zhen, H.; Chen, P.; Zheng, S.; Du, Y.; Chen, Z.; and Gan, C. 2023.
\newblock 3d-llm: Injecting the 3d world into large language models.
\newblock \emph{Advances in Neural Information Processing Systems}, 36:
  20482--20494.

\bibitem[{Huang et~al.(2024)Huang, Chen, Wang, Huang, Xu, Wang, Liu, Cheng,
  Zhao, Pang et~al.}]{huang2024chat}
Huang, H.; Chen, Y.; Wang, Z.; Huang, R.; Xu, R.; Wang, T.; Liu, L.; Cheng, X.;
  Zhao, Y.; Pang, J.; et~al. 2024.
\newblock Chat-scene: Bridging 3d scene and large language models with object
  identifiers.
\newblock \emph{Advances in Neural Information Processing Systems}, 37:
  113991--114017.

\bibitem[{Huang et~al.(2026)Huang, Wu, Xie, and Han}]{huang20263drs}
Huang, X.; Wu, J.; Xie, Q.; and Han, K. 2026.
\newblock 3drs: Mllms need 3d-aware representation supervision for scene
  understanding.
\newblock \emph{Advances in Neural Information Processing Systems}, 38:
  67961--67988.

\bibitem[{Jain et~al.(2022)Jain, Gkanatsios, Mediratta, and
  Fragkiadaki}]{jain2022bottom}
Jain, A.; Gkanatsios, N.; Mediratta, I.; and Fragkiadaki, K. 2022.
\newblock Bottom up top down detection transformers for language grounding in
  images and point clouds.
\newblock In \emph{European Conference on Computer Vision}, 417--433. Springer.

\bibitem[{Ju et~al.(2024)Ju, Sun, Du, Yuan, Ren, and Liu}]{ju2024large}
Ju, T.; Sun, W.; Du, W.; Yuan, X.; Ren, Z.; and Liu, G. 2024.
\newblock How large language models encode context knowledge? a layer-wise
  probing study.
\newblock In \emph{Proceedings of the 2024 Joint International Conference on
  Computational Linguistics, Language Resources and Evaluation (LREC-COLING
  2024)}, 8235--8246.

\bibitem[{Li et~al.(2023)Li, Li, Savarese, and Hoi}]{li2023blip}
Li, J.; Li, D.; Savarese, S.; and Hoi, S. 2023.
\newblock Blip-2: Bootstrapping language-image pre-training with frozen image
  encoders and large language models.
\newblock In \emph{International conference on machine learning}, 19730--19742.
  PMLR.

\bibitem[{Liu et~al.(2023)Liu, Li, Wu, and Lee}]{liu2023visual}
Liu, H.; Li, C.; Wu, Q.; and Lee, Y.~J. 2023.
\newblock Visual instruction tuning.
\newblock \emph{Advances in neural information processing systems}, 36:
  34892--34916.

\bibitem[{Liu, Qi, and Fu(2021)}]{liu20213d}
Liu, Z.; Qi, X.; and Fu, C.-W. 2021.
\newblock 3d-to-2d distillation for indoor scene parsing.
\newblock In \emph{Proceedings of the IEEE/CVF conference on computer vision
  and pattern recognition}, 4464--4474.

\bibitem[{Lopez-Paz et~al.(2015)Lopez-Paz, Bottou, Sch{\"o}lkopf, and
  Vapnik}]{lopez2015unifying}
Lopez-Paz, D.; Bottou, L.; Sch{\"o}lkopf, B.; and Vapnik, V. 2015.
\newblock Unifying distillation and privileged information.
\newblock \emph{arXiv preprint arXiv:1511.03643}.

\bibitem[{Luo et~al.(2018)Luo, Hsieh, Jiang, Niebles, and
  Fei-Fei}]{luo2018graph}
Luo, Z.; Hsieh, J.-T.; Jiang, L.; Niebles, J.~C.; and Fei-Fei, L. 2018.
\newblock Graph distillation for action detection with privileged modalities.
\newblock In \emph{Proceedings of the European conference on Computer Vision
  (ECCV)}, 166--183.

\bibitem[{Ma et~al.(2022)Ma, Yong, Zheng, Li, Liang, Zhu, and
  Huang}]{ma2022sqa3d}
Ma, X.; Yong, S.; Zheng, Z.; Li, Q.; Liang, Y.; Zhu, S.-C.; and Huang, S. 2022.
\newblock Sqa3d: Situated question answering in 3d scenes.
\newblock \emph{arXiv preprint arXiv:2210.07474}.

\bibitem[{Vapnik and Vashist(2009)}]{vapnik2009new}
Vapnik, V.; and Vashist, A. 2009.
\newblock A new learning paradigm: Learning using privileged information.
\newblock \emph{Neural networks}, 22(5-6): 544--557.

\bibitem[{Wu et~al.(2026)Wu, Liang, Feng, Xia, Zhang, Li, Tan, and
  Bai}]{wu2026generation}
Wu, X.; Liang, D.; Feng, T.; Xia, K.; Zhang, Y.; Li, X.; Tan, X.; and Bai, X.
  2026.
\newblock Generation Models Know Space: Unleashing Implicit 3D Priors for Scene
  Understanding.
\newblock \emph{arXiv preprint arXiv:2603.19235}.

\bibitem[{Yang et~al.(2025)Yang, Chen, Madaan, Iyengar, Qian, Fouhey, and
  Chai}]{yang20253d}
Yang, J.; Chen, X.; Madaan, N.; Iyengar, M.; Qian, S.; Fouhey, D.~F.; and Chai,
  J. 2025.
\newblock 3d-grand: A million-scale dataset for 3d-llms with better grounding
  and less hallucination.
\newblock In \emph{Proceedings of the Computer Vision and Pattern Recognition
  Conference}, 29501--29512.

\bibitem[{Yeh et~al.(2026)Yeh, Qian, Wang, Ma, Tighe, and Xiao}]{yeh2026beyond}
Yeh, C.-H.; Qian, S.; Wang, M.; Ma, Y.; Tighe, J.; and Xiao, F. 2026.
\newblock Beyond 3D VQAs: Injecting 3D Spatial Priors into Vision-Language
  Models for Enhanced Geometric Reasoning.
\newblock In \emph{Proceedings of the IEEE/CVF Conference on Computer Vision
  and Pattern Recognition}, 16723--16733.

\bibitem[{Zhang et~al.(2026)Zhang, Zhou, Liu, Kadambi, and
  Fan}]{zhang2026spatialstack}
Zhang, J.; Zhou, S.; Liu, B.; Kadambi, A.; and Fan, Z. 2026.
\newblock Spatialstack: Layered geometry-language fusion for 3d vlm spatial
  reasoning.
\newblock In \emph{Proceedings of the IEEE/CVF Conference on Computer Vision
  and Pattern Recognition}, 38678--38688.

\bibitem[{Zhang et~al.(2024)Zhang, Wu, Li, Li, Ma, Liu, and
  Li}]{zhang2024llava}
Zhang, Y.; Wu, J.; Li, W.; Li, B.; Ma, Z.; Liu, Z.; and Li, C. 2024.
\newblock Llava-video: Video instruction tuning with synthetic data.
\newblock \emph{arXiv preprint arXiv:2410.02713}.

\bibitem[{Zheng, Huang, and Wang(2025)}]{zheng2025video}
Zheng, D.; Huang, S.; and Wang, L. 2025.
\newblock Video-3d llm: Learning position-aware video representation for 3d
  scene understanding.
\newblock In \emph{Proceedings of the IEEE/CVF Conference on Computer Vision
  and Pattern Recognition}, 8995--9006.

\bibitem[{Zhu et~al.(2025)Zhu, Wang, Zhang, Pang, and Liu}]{zhu2025llava}
Zhu, C.; Wang, T.; Zhang, W.; Pang, J.; and Liu, X. 2025.
\newblock Llava-3d: A simple yet effective pathway to empowering lmms with 3d
  capabilities.
\newblock In \emph{Proceedings of the IEEE/CVF International Conference on
  Computer Vision}, 4295--4305.

\bibitem[{Zhu et~al.(2023)Zhu, Ma, Chen, Deng, Huang, and Li}]{zhu20233d}
Zhu, Z.; Ma, X.; Chen, Y.; Deng, Z.; Huang, S.; and Li, Q. 2023.
\newblock 3d-vista: Pre-trained transformer for 3d vision and text alignment.
\newblock In \emph{Proceedings of the IEEE/CVF International Conference on
  Computer Vision}, 2911--2921.

\end{thebibliography}
\end{document}